\documentclass{article}

\usepackage{arxiv}

\usepackage[utf8]{inputenc}
\usepackage[T1]{fontenc}
\usepackage{hyperref}
\usepackage{url}
\usepackage{booktabs}
\usepackage{amsfonts}
\usepackage{nicefrac}
\usepackage{microtype}
\usepackage{graphicx}
\usepackage{natbib}
\usepackage{doi}

\title{A rubric landscape for evaluating clinical reasoning in large language models: what exists, what is missing, and what needs to be combined}

\author{
  Zhangshu Joshua Jiang \\
  DRIVE-Health CDT \\
  Department of Biostatistics and Health Informatics \\
  Institute of Psychiatry, Psychology and Neuroscience \\
  King's College London \\
  London, UK \\
    Cleveland Clinic London \\
  London, UK \\
    \texttt{zhangshu.j.jiang@kcl.ac.uk}
  \And
   Zina Ibrahim \\
  DRIVE-Health CDT \\
  Department of Biostatistics and Health Informatics \\
  Institute of Psychiatry, Psychology and Neuroscience \\
  King's College London \\
  London, UK \\
    \And
   James Teo \\
  DRIVE-Health CDT \\
  Department of Biostatistics and Health Informatics \\
  Institute of Psychiatry, Psychology and Neuroscience \\
  King's College London \\
  London, UK \\
}

\renewcommand{\shorttitle}{Rubric landscape for clinical reasoning in LLMs}

\hypersetup{
pdftitle={A rubric landscape for evaluating clinical reasoning in large language models: what exists, what is missing, and what needs to be combined},
pdfsubject={cs.CL, cs.AI},
pdfauthor={Zhangshu Joshua Jiang},
pdfkeywords={clinical reasoning, large language models, evaluation, rubrics, benchmarks, LLM-as-judge, electronic health records, faithfulness},
}

\date{First drafted 26 August 2026; revised 1 October 2026}

\begin{document}
\maketitle

\begin{abstract}
\textbf{Background.} Large language models (LLMs) are increasingly evaluated on clinical tasks, but exam-style accuracy says little about whether a model reasons well over a patient's record. Clinical reasoning, operationalised here as the structured process of integrating and updating information across time and evidence sources to form, revise and justify a representation of a patient's problem and a defensible plan, is a different construct from answer correctness, and it is not obvious that any existing instrument measures it.

\textbf{Aim.} To map the rubrics and benchmarks that could be used to score clinical reasoning in LLM outputs, to state precisely which components of the construct each one measures, and to identify what must be combined or newly built to evaluate reasoning over longitudinal, multi-document clinical records.

\textbf{Approach.} A structured narrative review of three literatures: medical education instruments developed to assess human learners (IDEA, R-IDEA, ART, the Script Concordance Test); clinical LLM benchmarks published from 2023 onwards (including HealthBench, MedR-Bench, TIMER-Eval, ER-Reason, SCT-Bench, MedThink-Bench and recent uncertainty, counterfactual and omission benchmarks); and general-domain evaluation methodology for long-form generation (LLM-as-judge, checklist decomposition, importance-aware factuality, judge reliability audits). Preprints are included and flagged.

\textbf{Findings.} No single instrument captures the construct. Problem representation and differential or management reasoning
are reasonably covered, although reliability varies by instrument
and assessment setting. TIMER-Eval directly targets temporal synthesis, while ER-Reason assesses sequential diagnostic belief updating. Dedicated evaluations of uncertainty preservation and
counterfactual robustness are emerging, but their applicability
to longitudinal free-text reasoning remains limited. Factual completeness is well theorised in the general domain, where omission rather than fabrication is the dominant failure mode, with early clinical confirmation. Reasoning faithfulness remains the weakest dimension: we identified one clinical causal-ablation study on multiple-choice questions.

\textbf{Implications.} Existing tools should be used only for the parts they actually measure, and joined using five methodological commitments: binary compilation of rubric items, separation of completeness from correctness, case-specific importance weighting with non-compensable safety caps, order-consistency checks for the temporal axis, and chance-corrected reliability reporting throughout. Genuine design work remains for calibrated uncertainty, counterfactual reasoning and faithfulness over longitudinal free-text records.
\end{abstract}


\section{Introduction}
Large language models are now routinely evaluated on medical tasks, and the dominant instruments remain multiple-choice examinations and short-answer question sets. These are automatable and reproducible, but their construct validity for clinical work is poor: Alaa and colleagues show that, for a frontier model, correctness on exam items only weakly predicts correctness on real clinical cases, and that model rankings can reverse when the same models are assessed on cases derived from health records \citep{alaa2025}. A benchmark optimised for exam accuracy may therefore rank models in the wrong order for the tasks that matter clinically. The natural response is to score the reasoning rather than the answer. That response immediately raises two problems: what exactly is being scored, and whether any existing rubric or benchmark scores it.

The first problem is definitional. There is no agreed definition of clinical reasoning in the medical education literature, and this is itself a settled finding rather than a gap in reading \citep{young2018}. The work reviewed here therefore does not attempt a general definition. It commits to a narrower operational one: clinical reasoning is the process by which a reasoner integratesand updates information across time and evidence sources to form, revise and justify representations of a patient's clinical problem, and to generate defensible decisions about further investigation and management. That definition is deliberately scoped to reasoning from documented evidence, and in particular to reasoning over longitudinal, multi-document electronic health records in which later entries may legitimately supersede, rather than merely supplement, earlier ones. 

A caveat on what is being assessed. The dimensions below are a working way of organising the problem, not an established model of how clinicians think, and they overlap. For LLM outputs, this review treats the target as \emph{expressed} clinical reasoning: what a response shows under a defined task and evidence boundary, not whether the explanation reflects how the model produced it. Internal-process faithfulness is a related but separate question, discussed in Section~\ref{sec:faithfulness}. Whether this split into six dimensions is useful is an empirical question, not something this review can settle.

It is organised here into six provisional dimensions: problem representation; temporal or longitudinal synthesis; counterfactual reasoning; differential and management reasoning; calibrated uncertainty; and the faithfulness of stated reasoning to the process that actually produced the output. Whether stated reasoning is faithful is included because, in the LLM setting, a coherent rationale cannot be assumed to be an account of how the answer was generated \citep{turpin2023}.

The second problem, whether an existing instrument measures this construct, is the subject of this paper. The question is narrower than ``how good are medical LLMs'' and more practical than a taxonomy of evaluation methods. It asks, for each candidate rubric or benchmark, what it actually measures, what it does not, and what would need to be added or combined to score reasoning over serial clinical records. The answer matters for anyone planning to evaluate or fine-tune models on longitudinal clinical text, because the tools most often cited for this purpose were built for other jobs, and their labels do not always match their contents.

\section{Scope and approach}
This is a structured narrative review rather than a systematic review. Sources were identified through PubMed, the ACL Anthology, arXiv and Semantic Scholar, with forward and backward citation chasing from the core instruments, and screened by a single reviewer. The review considers LLM evaluation work published from 2023
onwards, with no lower date bound for medical education instruments. Preprints are included because the LLM evaluation literature moves largely through them; each is flagged in the text and the reference list, and findings from preprints are treated as provisional. No quantitative synthesis is attempted. Source selection was purposive and iterative rather than exhaustive. A complete search log was not retained, limiting reproducibility. The review is therefore best read as a reasoned map of the field against a stated construct, not as an exhaustive inventory.

Three adjacent literatures were considered but are not the focus of this review. The clinical decision support systems literature evaluates whether tools change clinical behaviour or outcomes, not the quality of reasoning steps, and is therefore out of scope here \citep{wright2008}. The verbal protocol and think-aloud literature from cognitive science underpins much of clinical reasoning theory, but it provides a data collection method rather than a scoreable rubric, and is not designed for post-hoc assessment of static text outputs \citep{fonteyn1993,pinnock2015}. The diagnostic error and patient safety literature offers retrospective tools for categorising errors after the fact, but is concerned with outcome rather than reasoning process \citep{ahrq,singh2015}. The remainder of this review therefore focuses on the traditions that most directly bear on the evaluation task at hand.

\section{The landscape in outline}
There is not currently one single rubric or benchmark in the literature reviewed here that captures the version of clinical reasoning defined above \citep{baker2015,schaye2022,thammasitboon2018,lubarsky2013,arora2025,qiu2025,cui2025,mehandru2025,gao2023,jeong2024,du2026,adewuyi2026,oukelmoun2025,meincke2026}. Instead, the literature falls into three fairly separate strands: medical education tools developed to assess human learners, newer benchmarks developed to assess large language model outputs, and a general-domain evaluation-methodology literature that determines whether any rubric can be applied reliably at scale \citep{liu2023,zheng2023,kim2024,ye2024,min2023,song2024,wei2024,lee2025,ruan2025,jafari2026,chen2026,wanner2025,reliability2026,rao2026autorubric,jacovi2020}. These strands overlap to some extent, but they are not doing the same job. Each is taken in turn below, followed by a dimension-by-dimension account of what is covered, partially covered and missing, and a statement of what needs to be combined.

\section{The medical education strand}
The genealogy of the most directly useful tools begins with the original IDEA assessment tool, developed by Baker and colleagues to score clinical reasoning documentation across four domains: interpretive summary, differential diagnosis, explanation of reasoning, and alternatives \citep{baker2015}. IDEA is worth citing in its own right, and not only as the ancestor of its revision, because its validity evidence already documents the central problem this review is concerned with. The original IDEA studies reported moderate inter-rater reliability for reporting, diagnostic reasoning and decision-making scores
(ICCs 0.53, 0.64 and 0.63, respectively), supporting formative rather than high-stakes use \citep{baker2015}. These are study-specific estimates, not an upper bound on reasoning-assessment reliability.
Revised-IDEA subsequently reported an ICC of 0.84 (95\% CI 0.74--0.90) \citep{schaye2022}.

That fragility is precisely what motivated Revised-IDEA, referred to here as R-IDEA. Schaye and colleagues redesigned the instrument to assess clinical reasoning documentation in admission notes with improved reliability, retaining the four-part structure of interpretive summary, differential diagnosis, explanation, and alternatives \citep{schaye2022}. In practice, what R-IDEA measures is how well reasoning is written down in a clinical note. That makes it useful for problem representation and parts of differential reasoning, but it does not assess temporal reasoning across serial records, calibrated uncertainty, counterfactual reasoning, or whether the written explanation actually drove the answer.

ART is useful for a slightly different reason. It was designed as a theory-informed tool for assessing diagnostic reasoning and feedback, and includes areas such as problem representation, prioritised differential reasoning, and metacognition \citep{thammasitboon2018}. Its value here is not that it solves model evaluation, but that it shows how medical educators try to break reasoning into visible parts that can be discussed and judged. Even so, ART still assesses human performance in presentation-based settings, and its metacognitive elements only really work if the speaker is assumed to be sincerely describing their own uncertainty or revision of thought.

The Script Concordance Test (SCT) is narrower again. It looks at how new information changes the plausibility of a hypothesis under uncertainty by comparing responses with those of an expert panel \citep{lubarsky2013}. That makes it relevant to structured judgement under uncertainty, but it has not traditionally been used as a general rubric for open-ended clinical reasoning text. The paradigm has recently been scaled for model evaluation: McCoy and colleagues assembled a public benchmark of 750 validated SCT items drawn from ten international datasets and compared ten state-of-the-art models against a human baseline of students, residents, and attending physicians \citep{mccoy2025}. Performance tracked clinical expertise, yet even reasoning-optimised models scored substantially lower on SCT than on conventional multiple-choice examinations, and response-pattern analysis showed systematic overconfidence, with models disproportionately favouring extreme belief shifts (+2 or $-$2) and underusing neutral ratings \citep{mccoy2025}. This suggests that chain-of-thought optimisation may encourage premature commitment rather than flexible probabilistic updating. Scaled SCT therefore exposes gaps in belief revision under uncertainty, but remains a deliberately constrained proxy: it supports a separate belief-updating task and supplies convergent evidence, but does not convert free-text uncertainty handling into an SCT score.

Taken together, the medical education literature is strongest when the task is to assess how a human learner frames a problem, prioritises a differential, and justifies a plan \citep{baker2015,schaye2022,thammasitboon2018,lubarsky2013}.

Three broader points from the education literature shape how this review frames the task. Problem representation is usually described as an abstraction of the case into clinically meaningful features that supports hypothesis generation \citep{bowen2006}. Management reasoning is increasingly treated as distinct from diagnostic reasoning, often with more than one defensible plan \citep{cook2019}. And assessment of clinical reasoning requires an explicit construct, task format, scoring approach and justification for how scores are interpreted \citep{gordon2022}. These points suggest that problem representation is better treated as one contribution to a response judged across several criteria than as a stand-alone target.

Its main limitation for the present purpose is structural: these tools generally assume that what someone writes or says can be treated as a meaningful account of their reasoning \citep{baker2015,schaye2022,thammasitboon2018,lubarsky2013}. The IDEA validity work shows that this assumption was already under strain for human authors \citep{baker2015}; it becomes much shakier still when the object being assessed is an LLM output rather than a trainee note or presentation.

\section{The clinical LLM benchmark strand}
The recent LLM evaluation literature starts from a different place \citep{arora2025,qiu2025,cui2025,mehandru2025,gao2023,jeong2024,du2026,adewuyi2026,oukelmoun2025,meincke2026}. Most of these benchmarks are designed to judge the quality, safety, completeness, or usefulness of a model's answer, rather than to assess clinical reasoning in a mechanism-based reasoning used in medical education. HealthBench is a good example. It uses physician-authored rubrics to score healthcare conversations across dimensions such as accuracy, completeness, communication, context-awareness, and instruction-following, with individually weighted criteria that can carry negative weights for harmful content \citep{arora2025}. What it gives you, then, is a structured way of judging clinically relevant answers, and its weighting machinery is a useful precedent for the importance-weighted approach set out below. What it does not obviously give you is a full rubric for clinical reasoning in the narrower sense defined above. A correct final answer does not show that the supporting reasoning was sound, which is why the reasoning-focused work below matters \citep{rao2026seq}.

MedR-Bench gets closer to reasoning-focused evaluation. It uses 1,453 structured patient cases with reference reasoning derived from case reports, comprising 957 diagnosis cases and 496 treatment-planning cases, including a substantial rare-disease subset \citep{qiu2025}. That makes it more informative than short-answer medical question-answering benchmarks, because it compares generated reasoning against structured case material rather than only checking the final answer. Even so, it still assesses correspondence to reference reasoning and properties of the written rationale, not whether the stated reasoning steps actually caused the answer. For that reason, it is best treated here as a partial proxy for reasoning quality rather than as a solution to the faithfulness problem.

TIMER is the clearest example of a benchmark adding something distinct, though it needs to be characterised precisely. TIMER combines temporally grounded instruction tuning with an evaluation schema, TIMER-Eval, for longitudinal clinical records. Instruction-response pairs are linked to timestamps in patient records, while TIMER-Eval assesses questions requiring synthesis across multiple timepoints \citep{cui2025}. The evaluation component looks at whether model outputs respect temporal boundaries, identify trends correctly, and preserve chronological order. This matters because temporal reasoning is often mentioned in general terms in the clinical reasoning literature, but is rarely measured directly in a scored tool. TIMER-Eval therefore looks genuinely useful for the temporal component of a clinical reasoning rubric. Two limits should be stated. First, it measures temporal reasoning performance, not the whole of clinical reasoning. Second, the temporal distribution of evaluation items requires attention: the earlier MedAlign benchmark is heavily recency-weighted, with 55.3\% of instructions referring to the final quarter of the timeline, and TIMER-Eval addresses this by explicitly providing Edge and Uniform temporal distributions \citep{cui2025}. Even so, it does not explicitly test whether a model recognises that a later record supersedes, rather than merely supplements, an earlier one. That distinction, between accumulation and legitimate supersession, is a defining feature of serial clinical records, and it remains untested even after TIMER.

ER-Reason grounds evaluation in real emergency department
documentation and operationalises clinical reasoning as sequential
belief updating. The current version comprises 25,174 de-identified
notes from 3,437 patients across 3,984 encounters. Its reasoning
evaluation includes 194 physician-authored cases containing 787
SCT-style items, with 2,555 physician annotations. These assess
whether models update diagnostic beliefs in the appropriate
direction and magnitude as evidence accumulates \citep{mehandru2025}.
However, these structured tasks do not establish the faithfulness
of unrestricted free-text reasoning or provide a general rubric
for multi-document outputs.

This sequential-workflow paradigm has been extended and validated against clinician cohorts in several 2026 studies. Brodeur and colleagues tested an advanced model against a baseline of several hundred physicians across five clinical reasoning experiments (differential diagnosis generation, display of reasoning, triage differential, probabilistic reasoning, and management reasoning), adjudicated by expert physicians with validated psychometrics, and complemented these with a real-world emergency department trial comparing human and model second opinions at a tertiary academic centre \citep{brodeur2026}. The model matched or exceeded physician performance on text-based triage and reasoning under uncertainty, but remained bounded by its text-only input and absence of bedside multimodal cues \citep{brodeur2026}. That boundary applies equally to any evaluation, including the one motivating this review, that scores reasoning from text alone.

Rao and colleagues expose the vulnerability that sits underneath such aggregate performance. Evaluating 21 frontier models across identical sequential workflow stages on 29 standardised vignettes, they found performance strongest at final diagnosis and management and weakest at early diagnostic reasoning and differential generation, and characterised a consistent failure mode of premature diagnostic closure: convergence on a single diagnosis when uncertainty and competing hypotheses should have been sustained \citep{rao2026seq}. High accuracy on final diagnostic targets therefore masks intermediate reasoning failure, which is the clearest available empirical argument for scoring trajectories rather than endpoints.

To move beyond final-outcome matching and capture these intermediate reasoning errors, recent work has pivoted toward highly structured, expert-authored rubrics. Ismail and colleagues addressed the limitation of unweighted checklists by introducing five deliberately difficult clinician-authored scenarios, each accompanied by an atomic, weighted, mutually exclusive and collectively exhaustive rubric comprising 25 to 62 criteria per task (184 criteria in total) \citep{ismail2026}. Their controlled comparison of frontier language models revealed what they term an inversion of clinical priority: while models scored between 80\% and 90\% on low-stakes, simple checklist items, they satisfied critical, safety-essential (weight-5) criteria at rates of only 32.4\% to 41.7\%, and over half of these critical criteria were met by no model \citep{ismail2026}. This indicates that evaluations which ignore clinical risk weightings fail to detect dangerous clinical omissions.

MedThink-Bench addresses scalability of trajectory scoring. Comprising 500 complex medical question and answer pairs across ten domains annotated with expert-crafted step-by-step rationales, it scores intermediate ``scoring points'' rather than final answers, and its LLM-w-Ref evaluation framework achieves strong correlation with expert clinician judgement \citep{zhou2025}. Read alongside the judge-reliability audit discussed below \citep{reliability2026}, this correlation should be treated as promising rather than settled, since correlation with experts is not the same as chance-corrected agreement.

DR.BENCH is helpful mainly as a warning. It packages six tasks under the heading of diagnostic reasoning, but scores them using generation metrics such as ROUGE-L and F1 \citep{gao2023}. That means it should be read mainly as a benchmark of task performance under those metrics, not as a strong measure of reasoning quality in a broader sense. In this review, it is therefore more useful as a reminder not to take benchmark labels at face value than as a model for a reasoning rubric. OLAPH, which improves and evaluates factuality in biomedical long-form question answering, sits adjacent to this strand and is noted here as the main clinical entry point into the long-form factuality literature discussed below \citep{jeong2024}.

Three recent additions to this strand bear directly on the gaps this review identifies, and each partially fills one of them. Du and colleagues operationalise diagnostic uncertainty preservation as a scoreable property: across 1,200 clinical documents annotated with over nine thousand propositions on a five-level uncertainty scale, they find that models preserve the original diagnostic uncertainty level in under half of cases, collapsing ``possible'' findings into definite ones or hedging definite findings into vague ones \citep{du2026}. This is the first tool in the checked sources that treats preservation of expressed diagnostic uncertainty as a measurable dimension rather than a desideratum, and its five-level taxonomy is directly reusable for rubric item design. MamaBench does analogous work for counterfactual reasoning: it constructs matched counterfactual pairs of clinical narratives in maternal and child health (434 narratives forming 217 pairs) in which a single discriminative clinical parameter changes the correct diagnosis, and measures whether models change their answer accordingly, reporting a ``Bias Trap Rate'' and identifying a characteristic failure mode of diagnostic fixation on the unperturbed presentation \citep{adewuyi2026}. It is narrow in specialty and works on paired vignettes rather than longitudinal free text, but it demonstrates that counterfactual clinical robustness can be operationalised as a scored quantity. Third, Oukelmoun and colleagues treat omission detection in LLM-generated medical summaries as a problem distinct from hallucination detection, with a dedicated detection method \citep{oukelmoun2025}; this is the only paper found in the checked sources that separates the two failure modes explicitly in clinical text, and a natural baseline for any rubric-based completeness metric. A working paper by Meincke and colleagues, which frames diagnosis in a high-fidelity patient simulation as a partially observable sequential decision problem and reports accuracy conditional on the model's stated confidence, provides rare empirical calibration data in a clinically realistic setting, though it is not yet peer-reviewed and is cited here with that caveat \citep{meincke2026}.

\section{Evaluation methodology as a separate literature}
A third body of work sits alongside the medical education and clinical benchmark strands: general-domain research on how to score long-form generation reliably. This literature is not clinical, but it is methodologically decisive for any clinical rubric, because it determines whether that rubric can be applied at scale.

Its genealogy is worth a sentence, because it explains why the field converged on checklists. The first generation of LLM-as-judge evaluation used holistic judgements; G-Eval's chain-of-thought scoring \citep{liu2023} and the MT-Bench and Chatbot Arena validity study \citep{zheng2023} are the founding examples. The move to explicit, fine-grained criteria came with Prometheus \citep{kim2024} and FLASK \citep{ye2024}, which decompose judgement into named skills or rubric dimensions. In parallel, factuality evaluation developed a decompose-search-verify pipeline (FActScore \citep{min2023}, VeriScore \citep{song2024} and SAFE \citep{wei2024}) which scores the precision of atomic claims but is structurally silent on whether a response contains the claims it should have contained. Both lineages feed directly into the tools a clinical reasoning rubric must rely on.

Two findings from the checklist literature are directly consequential. Lee and colleagues (CheckEval) show that LLM-as-judge unreliability derives substantially from combining subjective criteria with Likert-scale scoring; decomposing evaluation into binary checklist questions raised average inter-evaluator agreement across twelve judge models by 0.45 and reduced score variance, while retaining strong correlation with human judgement \citep{lee2025}. One qualification should be attached to that figure: it is raw agreement, not chance-corrected reliability. A recent large-scale audit of twenty-one judge models across roughly half a million judgements found that raw agreement systematically overstates chance-corrected kappa by 33 to 41 percentage points, that judge rankings shift substantially across benchmarks, and that high test-retest consistency can coexist with severe position bias \citep{reliability2026}. This does not invalidate the CheckEval result, but it does mean that any rubric built on binary items and LLM judges, including the one motivating this review, must report chance-corrected agreement and test for positional and formatting biases, rather than treating raw agreement as evidence of judge trustworthiness.

The second consequential finding comes from Ruan and colleagues (ExpertLongBench and the CLEAR framework), who evaluate expert-level long-form generation across eleven tasks in nine domains including medicine, scoring outputs against expert-authored structured checklists. They found that models covered over two-thirds of expert-rubric checklist items on average, while the best-performing model achieved an overall task F1 of only 33.4, and that checklist coverage was only weakly correlated with output accuracy \citep{ruan2025}. Their conclusion, that high coverage can make incorrect output appear expert-aligned, is the central methodological warning for clinical rubric design.

A convergent third strand concerns claim importance and completeness. Jafari and colleagues argue that existing factuality evaluation is overwhelmingly precision-oriented (responses are decomposed into atomic claims and each is verified), which says nothing about whether the response covers the facts it should have included; they propose measuring precision and recall jointly, weighting reference facts by relevance and salience, and find that current models perform markedly better on precision than recall \citep{jafari2026}. Chen and colleagues (GAMUT) reach a convergent conclusion from a different direction with a two-level meta-rubric: a structured expert-authored representation that encodes content organisation and importance tiers, mechanically compiled to a flat checklist of binary, LLM-gradable items, with contradiction penalised more heavily than omission \citep{chen2026}. Across fourteen frontier models on their benchmark, they report that omission, rather than fabrication, is the dominant failure mode in long-form generation, accounting for roughly two-thirds of failures in the weakest models and over a quarter even in the strongest \citep{chen2026}. One qualification is required and is stated plainly: GAMUT is a general-domain, multimodal benchmark built on consumer imagery, in which medical content amounts to roughly one percent of the dataset and consists of consumer-facing factual lookups rather than clinical reasoning \citep{chen2026}. Its omission-dominance finding is therefore adopted here as a general-domain result extrapolated to the clinical case, not as clinical evidence. The extrapolation is nonetheless defensible, because the same signature is already visible inside the clinical literature: MedR-Bench reports high step-level accuracy alongside omissions at critical reasoning steps \citep{qiu2025}, and dedicated clinical omission-detection work confirms that omission is a distinct and prevalent failure mode in medical summarisation \citep{oukelmoun2025}. Wanner and colleagues (VITAL) independently show that treating all claims as equally important obscures errors in key information \citep{wanner2025}. The clinical analogue is obvious: the unmentioned red-flag diagnosis or unacknowledged contraindication is a recognised source of harm.

Five design commitments inform the proposed instrument.
These are design recommendations rather than implemented
or validated features. First, we propose defining rubric dimensions with human-facing anchors for construct clarity and compiling them into binary items for automated scoring, a separation supported by CheckEval and GAMUT's meta-rubric architecture \citep{lee2025,chen2026}. Second, completeness and correctness are scored as separate quantities rather than combined into a single quality judgement, since high coverage can mask error \citep{ruan2025,jafari2026}; on this point the approach deliberately departs from GAMUT, which folds both into one asymmetrically weighted score \citep{chen2026}. Third, reference facts are converted into case-specific required, optional and prohibited items with pre-specified clinical-importance weights, so that omission of a safety-critical item is not equivalent to omission of a minor detail; this can be operationalised with an existing framework such as AutoRubric, which supports weighted criteria, negative weights for unsafe content, ensemble judging, and psychometric reliability reporting \citep{rao2026autorubric}, and has a clinical precedent in HealthBench's negatively weighted harm criteria \citep{arora2025}. Fourth, in light of the judge-reliability audit, all automated scoring should report chance-corrected agreement statistics and position-bias checks, not raw agreement alone \citep{reliability2026}.

Fifth, the temporal axis should include order-consistency checks, testing whether a response preserves the sequence and status of events, including whether later documentation supersedes earlier information. This is a design proposal, not an implemented feature.

Two limitations of this literature should be stated plainly. None of these frameworks evaluates whether stated reasoning caused the output, so the faithfulness gap identified below is not resolved by methodological refinement alone. And none accommodates longitudinal, multi-document input in which later records legitimately supersede earlier ones, a defining feature of serial clinical records. This limitation is particularly relevant to longitudinal clinical
records, where chronological ordering and the interpretation of later
information are integral to the meaning of a clinical narrative.
A temporal component of a rubric should therefore test whether
models preserve the order and status of clinically relevant events,
including whether later documentation supersedes earlier information.

\section{What is covered, what is partially covered, and what is missing}
Looking across the strands, the overall picture is more textured than a simple covered/uncovered division, and Table~\ref{tab:coverage} summarises it. Problem representation and differential or management reasoning are relatively well covered, at least in part \citep{baker2015,schaye2022,thammasitboon2018,lubarsky2013,arora2025}. Temporal reasoning is directly measured mainly by TIMER-Eval, with the supersession problem still open \citep{cui2025}. Calibrated uncertainty and counterfactual reasoning are better described as emerging: each now has at least one dedicated clinical operationalisation \citep{du2026,adewuyi2026}, but neither has a reusable rubric for longitudinal free-text outputs, and both existing tools work on static or paired vignettes rather than serial records. Factual completeness has moved from invisible to well-theorised in the general domain \citep{jafari2026,chen2026,wanner2025}, with early clinical instantiations \citep{oukelmoun2025}. Reasoning faithfulness remains the most weakly covered dimension, but the honest statement of that gap has changed, as follows.

\begin{table}[htbp]
\centering
\small
\caption{Coverage of the six clinical-reasoning dimensions by existing instruments and benchmarks.}
\label{tab:coverage}
\begin{tabular}{@{}p{2.6cm}p{3.4cm}p{4.3cm}p{4.3cm}@{}}
\toprule
\textbf{Dimension} & \textbf{Best-available instrument(s)} & \textbf{What is measured} & \textbf{What remains open} \\
\midrule
Problem representation & IDEA, R-IDEA, ART \citep{baker2015,schaye2022,thammasitboon2018} & Structured write-up of interpretive summary and differential in clinical notes & Reliability varies by instrument and setting; transfer to LLM
outputs requires validation; documentation quality does not
establish causal faithfulness \citep{baker2015,schaye2022} \\
Differential and management reasoning & HealthBench, MedR-Bench, ER-Reason, Brodeur et al. \citep{arora2025,qiu2025,mehandru2025,brodeur2026} & Quality, safety and completeness of clinically relevant answers; correspondence to reference reasoning & Correspondence to reference text is not causal reasoning; text-only input, no bedside cues \citep{brodeur2026} \\
Temporal / longitudinal synthesis & TIMER-Eval \citep{cui2025} & Respect for temporal boundaries, trend identification, chronological order & Does not test whether a later record supersedes, rather than supplements, an earlier one \\
Counterfactual reasoning & MamaBench \citep{adewuyi2026} & Bias Trap Rate on paired vignettes where one parameter changes the correct diagnosis & Single specialty; paired vignettes, not longitudinal free text \\
Calibrated uncertainty & Du et al. uncertainty-preservation benchmark, SCT-Bench \citep{du2026,mccoy2025} & Preservation of five-level uncertainty labels; belief-shift patterns against expert panels & No rubric for uncertainty that must change as records accumulate \\
Reasoning faithfulness & Afolabi et al. causal-ablation probe \citep{afolabi2025} & Causal necessity of individual chain-of-thought steps via redaction & Workshop-scale study on multiple-choice only; no free-text, longitudinal, rubric-integrated version \\
\bottomrule
\end{tabular}
\end{table}

\section{Reasoning faithfulness}
Reasoning faithfulness is the clearest gap, but it is no longer an empty one. The chain-of-thought literature shows that model explanations can be unfaithful to the processes that produce answers: Turpin and colleagues demonstrated that answers can be steered by biasing features the model never mentions in its explanation \citep{turpin2023}, and Arcuschin and colleagues extended this to realistic, unbiased prompts, documenting implicit post-hoc rationalisation at rates that vary widely across production models \citep{arcuschin2025}. A rationale that sounds coherent is therefore not necessarily a faithful account of how the output was generated. This matters because the medical education tools above were not designed to detect that problem \citep{baker2015,schaye2022,thammasitboon2018,lubarsky2013}. They generally assume that the explanation is part of the reasoning performance itself, whereas in the LLM setting that assumption needs to be tested rather than simply accepted.

There is now at least one direct clinical instantiation of that test. Afolabi and colleagues run a causal-ablation probe on medical question answering: reasoning steps are redacted one at a time, and a step is counted as causal only if its removal changes the model's prediction. They find that only around ten percent of chain-of-thought steps are individually causal, and that removing steps improved predictions about as often as it harmed them, a pattern consistent with post-hoc rationalisation rather than load-bearing reasoning \citep{afolabi2025}. This study is the closest existing partial solution to the faithfulness gap, and its redact-and-observe design is a template that a clinical reasoning rubric can adapt. Its limitations also delimit what remains open: it is a workshop-track study of three closed-source models on multiple-choice questions, and single-step redaction cannot distinguish an unnecessary step from one that is redundant with retained content \citep{afolabi2025}. No clinical work identified in this review performs causal mediation on internal representations or measures divergence between latent and verbalised reasoning; that sub-literature exists only in general-domain interpretability research. The gap is therefore precisely stated: not the absence of faithfulness testing in medicine, but the absence of a rubric-integrated, free-text, longitudinal version of it.

Jacovi and Goldberg provide the conceptual grounding here, defining faithfulness as the extent to which an explanation accurately reflects the reasoning process actually underlying an output, and arguing against a binary conception in favour of a graded one, on the grounds that a strict pass/fail standard is unrealistic and of limited practical use \citep{jacovi2020}. That graded framing is appropriate here. Faithfulness is better handled as a companion interventional study than as a domain scored from the output alone, using perturbation and counterfactual probing that extends the causal-ablation design from multiple-choice answers to free-text differential generation. Note that a change in output after a perturbation does not by itself show that the original explanation was faithful. One structural point deserves emphasis: faithfulness cannot be folded into the completeness or correctness axes, because a model can be complete and correct while its stated reasoning did not cause its answer. Checklist architectures verify what is stated, not why it was stated, so faithfulness requires interventional measurement by construction.

\section{Counterfactual reasoning}
Counterfactual reasoning is the second main gap, now partially narrowed. MamaBench demonstrates that counterfactual clinical robustness can be scored, and its paired-perturbation design and Bias Trap Rate metric are directly transferable, but it is confined to one specialty and to short paired vignettes \citep{adewuyi2026}. The checked sources still do not provide a rubric for scoring counterfactual clinical reasoning in longitudinal free-text outputs, where the perturbation of interest may be a changed result three documents back rather than a swapped parameter in a single vignette. The safer conclusion is therefore not that there is no related work, but that there is a transferable measurement design and no reusable tool for this specific task.

\section{Calibrated uncertainty}
Calibrated uncertainty follows the same pattern. Uncertainty preservation in clinical text is now measurable, and the finding that models preserve the source's diagnostic uncertainty level in under half of cases is itself a motivating result for treating this as a first-class rubric dimension \citep{du2026}. Early calibration data from simulated sequential diagnosis suggest that stated confidence does carry signal in clinically realistic settings \citep{meincke2026}. The characteristic failure mode is corroborated from two independent directions. Rao and colleagues document models collapsing complex differentials into premature diagnostic closure when competing possibilities are present \citep{rao2026seq}, and McCoy and colleagues show, against expert-panel distributions, a strong bias toward extreme SCT responses and near-avoidance of neutral ratings, evidence of an inability to make subtle probabilistic adjustments on new information \citep{mccoy2025}. What is still missing is the combination that longitudinal records require: calibration assessed over a case in which the appropriate uncertainty level itself changes as records accumulate.

\section{Factual completeness}
Whereas most benchmarks are precision-oriented and ask whether stated claims are correct \citep{min2023,song2024,wei2024}, the convergent findings of Jafari, GAMUT, and VITAL indicate that omission of required content is the more common and, clinically, the more dangerous failure \citep{jafari2026,chen2026,wanner2025}. This omission dominance has now received clinical confirmation. Ismail and colleagues showed that evaluating models using flat checklists or unweighted summaries masks critical clinical omissions: under a clinician-authored, weighted rubric, models achieved high coverage on low-stakes administrative or communication criteria while frequently omitting high-stakes, safety-critical actions such as emergency escalation or contraindicated-treatment warnings \citep{ismail2026}. This inversion of clinical priority indicates that completeness cannot be scored as a simple proportion of covered items; safety-critical omissions must act as non-compensable score caps. Zhou and colleagues add that checking intermediate step-level scoring points, rather than final-answer matching, is necessary to detect models that bypass critical logical steps \citep{zhou2025}.

\section{What needs to be combined}
The practical implication is fairly modest. An evaluation programme for clinical reasoning over serial records does not need to throw away existing tools; it needs to use them carefully and only for the parts they actually measure. R-IDEA provides a starting point for assessing written problem representation and differential reasoning, informed by the study-specific reliability findings for IDEA and Revised-IDEA \citep{baker2015,schaye2022}. Its reliability for assessing LLM outputs remains to be established. TIMER-Eval is the clearest starting point for assessing temporal reasoning over serial records, extended to test supersession rather than only chronology \citep{cui2025}. Uncertainty-preservation taxonomies \citep{du2026} and counterfactual pairing designs \citep{adewuyi2026} supply measurement patterns for two dimensions that previously had none, and the causal-ablation paradigm supplies one for faithfulness \citep{afolabi2025}. The rest of the rubric then needs to join the strands together: use the medical education literature to keep the rubric clinically meaningful, use the LLM benchmark literature for setting-specific measurement designs, and use the evaluation-methodology literature to make the whole thing psychometrically defensible at scale, through binary compilation for reliability \citep{lee2025}, separated completeness and correctness \citep{ruan2025,jafari2026}, importance weighting with safety caps \citep{arora2025,chen2026,wanner2025,rao2026autorubric}, order-consistency for the temporal axis, and chance-corrected reliability reporting throughout \citep{reliability2026}. Beyond that, genuine design work remains for the three dimensions where existing tools supply patterns but not instruments: calibrated uncertainty, counterfactual reasoning, and faithfulness over longitudinal free-text records.

Specifically, the integration draws on McCoy and colleagues' scaled concordance paradigm for structured judgement under uncertainty \citep{mccoy2025}, Brodeur and colleagues' multi-stage workflow design for mapping reasoning trajectories sequentially \citep{brodeur2026}, Rao and colleagues' premature-closure finding as the motivating failure mode for the differential axis \citep{rao2026seq}, and Ismail and colleagues' atomic weighted rubrics as the model for safety-critical prioritisation \citep{ismail2026}. These are proposed design commitments, not evidence that the resulting instrument is reliable or valid. Binary compilation and case-specific importance weighting remain planned build steps rather than implemented features \citep{lee2025,chen2026}, as does the specification of judge models, repeated runs, option-order randomisation, and acceptance thresholds. Findings from MedThink-Bench and Ismail and colleagues motivate testing automated grading \citep{zhou2025,ismail2026}, but agreement with independent clinician ratings must be established for the proposed instrument, using prespecified statistics and confidence intervals rather than correlation alone. This paper reports the design rationale for such an instrument, not a validated instrument.

\section{Limitations of this review}
Four limitations should be stated. First, screening and synthesis were performed by a single reviewer without dual screening or a formal flow diagram, so the review is a structured narrative map rather than a PRISMA-compliant scoping review. Second, the LLM evaluation literature moves largely through preprints, and several of the sources that fill the identified gaps (including the uncertainty-preservation, counterfactual, weighted-rubric and judge-reliability studies) had not completed peer review at the time of writing; their findings are treated as provisional and are flagged as such in the reference list. Third, the review is organised against one operational definition of clinical reasoning; instruments that look weak against these six dimensions may be well suited to other constructs, and the coverage judgements in Table~\ref{tab:coverage} should be read relative to the stated construct rather than as verdicts on the tools. Fourth, no quantitative synthesis was attempted, because the instruments reviewed do not share a common outcome, and reported figures (such as inter-rater reliabilities and agreement gains) are quoted from the primary sources without re-analysis.

\section{Conclusion}
No instrument identified in this review covers all six dimensions
of clinical reasoning over longitudinal records as defined here. The medical education tradition supplies clinically meaningful
anchors for problem representation and differential reasoning,
together with evidence that reliability depends on the instrument
and assessment setting. The clinical benchmark literature supplies setting-specific designs for temporal reasoning, sequential diagnostic updating, uncertainty preservation, counterfactual perturbation and safety-weighted completeness. The general evaluation-methodology literature supplies the machinery (binary compilation, separated completeness and correctness, importance weighting, order-consistency and chance-corrected reliability) that makes any of this applicable at scale. What remains to be built is the combination: a rubric that scores reasoning over serial records in which later entries supersede earlier ones, that treats calibrated uncertainty and counterfactual robustness as first-class dimensions, and that tests rather than assumes the faithfulness of stated reasoning. The concrete next step is not a larger benchmark but a small, carefully specified instrument whose reliability is established with chance-corrected statistics before any claim about model reasoning is made.

\section*{Declarations}
\noindent\textbf{Funding.} ZJJ is supported by the Engineering and Physical Sciences Research Council through the DRIVE-Health Centre for Doctoral Training at King's College London [EP/Y035216/1], with additional support from Cleveland Clinic London. The funders had no role in the preparation of the manuscript or the decision to submit it for publication.

\noindent\textbf{Competing interests.} The authors declare no competing interests.

\noindent\textbf{Data availability.} No new data were generated. All sources are cited in the reference list.

\noindent\textbf{Use of AI tools.} AI-based writing and research-assistance tools were used to support
literature synthesis and drafting. The authors reviewed and edited
the manuscript and take responsibility for its content, analysis
and conclusions.

\bibliographystyle{unsrtnat}
\bibliography{references}

\end{document}